\documentclass[a4paper, 10pt, conference]{ieeeconf}
\usepackage{FG2025}

\FGfinalcopy

\IEEEoverridecommandlockouts
\usepackage{graphicx}
\usepackage{amsmath}
\usepackage{amssymb}
\usepackage{booktabs}
\usepackage{algorithm,algpseudocode}
\usepackage{multicol}
\usepackage{multirow}
\usepackage{booktabs}
\usepackage[percent]{overpic}
\usepackage{xcolor,colortbl}
\let\labelindent\relax
\usepackage{enumitem}
\usepackage{pgfplots}
\usepackage{arydshln}
\usepackage[percent]{overpic}
\usepackage{caption}

\usepackage{fancyhdr}
\newcommand{\norm}[1]{\left\|#1\right\|}

\def\FGPaperID{XXX} 

\title{\LARGE \bf
GaussianGAN: Real-Time Photorealistic controllable Human Avatars
}

\author{\parbox{16cm}{\centering
    {\large Mohamed Ilyes Lakhal, Richard Bowden}\\
    {\normalsize
    CVSSP, University of Surrey, Guildford, United Kingdom}\\
    {\normalsize \texttt {\{m.lakhal, r.bowden\}@surrey.ac.uk}}}%
}

\begin{document}

\ifFGfinal
\thispagestyle{empty}
\pagestyle{empty}
\else
\author{Anonymous FG2025 submission\\ Paper ID \FGPaperID \\}
\pagestyle{plain}
\fi

\twocolumn[{%
\renewcommand\twocolumn[1][]{#1}%
\maketitle
\begin{center}
    \centering
    \captionsetup{type=figure}
    \includegraphics[width=.9\textwidth]{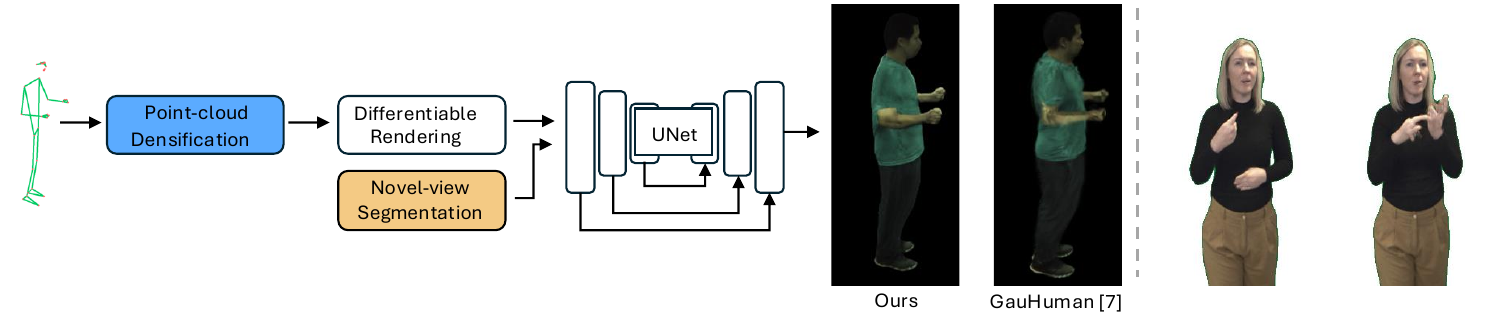}
    \captionof{figure}{GaussianGAN takes a $3$D human skeleton as input and then synthesises an image of a human from it. First, it reconstructs a dense point cloud by a novel densification strategy. Moreover, we propose a segmentation module that renders the semantics of the articulated object (\textit{e.g.} human) in the target view. We use a UNet encoder-decoder model to synthesise the human in motion. Our GaussianGAN can be easily applied to various problems, including animatable avatars and sign language production.}
    \label{fig:teaser}
\end{center}%
}]

 \thispagestyle{fancy}

\begin{abstract}

Photorealistic and controllable human avatars have gained popularity in the research community thanks to rapid advances in neural rendering, providing fast and realistic synthesis tools. However, a limitation of current solutions is the presence of noticeable blurring. To solve this problem, we propose GaussianGAN, an animatable avatar approach developed for photorealistic rendering of people in real-time. We introduce a novel Gaussian splatting densification strategy to build Gaussian points from the surface of cylindrical structures around estimated skeletal limbs. Given the camera calibration, we render an accurate semantic segmentation with our novel view segmentation module. Finally, a UNet generator uses the rendered Gaussian splatting features and the segmentation maps to create photorealistic digital avatars. Our method runs in real-time with a rendering speed of 79 FPS. It outperforms previous methods regarding visual perception and quality, achieving a state-of-the-art results in terms of a pixel fidelity of 32.94db on the ZJU Mocap dataset and 33.39db on the Thuman4 dataset.

\end{abstract}

\section{Introduction}
\label{sec:intro}
Animatable avatars~\cite{Peng_2021_CVPR,Peng_2021_ICCV} are becoming increasingly popular due to their wide range of applications in telepresence~\cite{Saito_2024_CVPR} and entertainment~\cite{weng_humannerf_2022_cvpr}. Traditional methods rely on expensive hardware acquisition systems that limit their applicability. Recent methods use a data-driven approach to reconstruct the avatar from monocular video, which enables application in the wild~\cite{hu2023gauhuman}.

Neural rendering techniques have emerged as the de facto paradigm for animatable avatars, as they enable photorealistic output while maintaining a reasonable frame rate during inference. Such methods rely on a human body template, such as the Skinned Multi-Person Linear Model (or SMPL), as a canonical space~\cite{Loper_2015_TOG}. In the NeRF-based methods, a backward flow is preferred, where each $3$D point is placed with respect to the canonical shape using the skinning procedure of SMPL~\cite{Yunus_CGF_2024}. More recent Gaussian splatting approaches use a forward flow to transfer each point representing the Gaussian parameters from the canonical to the observation space~\cite{Yunus_CGF_2024}. However, for applications with a lot of motion, such as sign language, SMPL estimation is error-prone due to depth ambiguity~\cite{Belhumeur_1997_CVPR,uy-scade-cvpr23} and hence affects the synthesis quality~\cite{Peng_2021_CVPR}. Therefore, some methods further refine the SMPL parameters to account for such inaccuracies~\cite{hu2023gauhuman,qian20233dgsavatar}.

\begin{figure*}[t!]
\begin{minipage}[t]{0.45\textwidth}
        \centering
        \vspace{-1.0em}
        \includegraphics[width=1.\textwidth]{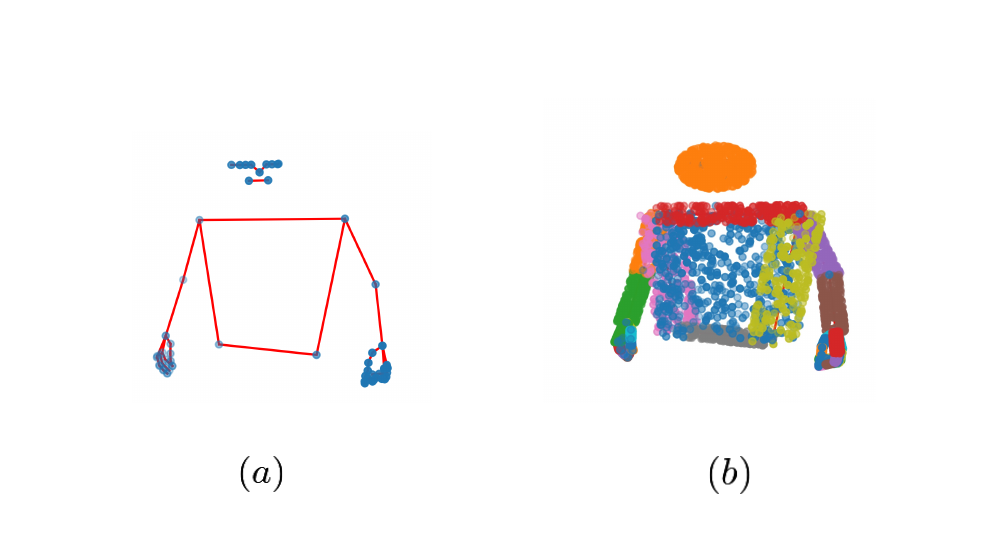} \vspace{-2em}
        \caption{Point cloud initialisation: (a) input skeleton; (b) output from our PCD densification (Alg.~\ref{alg:densify_pcd}).}
        \label{fig:densify_pcd}
    \end{minipage}
    \begin{minipage}[t]{0.5\textwidth}
\begin{algorithm}[H]
\footnotesize
\caption{Densify PCD}
\label{alg:densify_pcd}
  \begin{algorithmic}
    \State \textbf{Inputs:} $p = \{ p_i \}_{i=1}^J$: $3$D human skeleton of $J$ joints. $\mathcal{J} = \{ \mathbf{J}_i \}_{i=1}^L$: limbs connection dictionary.
    \State \textbf{Output:} $\mathcal{P} = \{ \mu_i \}_{i=1}^N$: Point cloud.\\
     
    \State $\mathcal{P} \gets \{ \}$
      
        \For{$\mathbf{J} \in \mathcal{J}$}
            \State $p_i, p_j \gets \mathbf{J}$
            \State $\mathbf{r} \gets 0.1 * \norm{ p_i - p_j }_2$
            \State $\mathcal{P} \gets \mathcal{P} \cup \texttt{build$\_$cylinder}(\mathbf{J}, \mathbf{r}) $
            
        \EndFor

    \end{algorithmic}
\end{algorithm}
\end{minipage}%
\end{figure*}

This paper presents GaussianGAN, a real time, photorealistic, controllable human avatar. Our GaussianGAN solves the blurring problem of the current state-of-the-art by providing a two-stage pipeline that can be trained end-to-end. The first stage leverages 3D Gaussian splatting to learn discriminative features of the posed human, which can be rendered (or splatted) into any view given camera calibration using differentiable rasterisation~\cite{Kato2020DifferentiableRA}. To achieve this, we create a point cloud from a 3D human skeleton to initialise a rough scene geometry. Then, an MLP network regresses the residuals of the Gaussian parameters to correct the actual body shape. In addition, we propose a novel-view semantic segmentation module that, following the task, renders segmentation maps to the image plane of the target view given calibration. This is an important part of the pipeline as descriptive and accurate modalities are essential for CNN-based generators~\cite {men2020controllable}. Finally, both the rendered Gaussian features and the segmentation maps are passed to a UNet~\cite{Xiaomeng_2018_TMI} generator to synthesise photorealistic images of the posed human in real time. This architecture was chosen because it can retain important spatial information through skip connections.

We summarise the contributions of this work as follows:
\begin{itemize}[noitemsep,topsep=0pt]
    \item[$\bullet$] \textbf{Densify PCD.} A densification strategy that builds the initial point cloud of Gaussian splatting parameters. In particular, we use the bone structure of the human skeleton to construct cylinders for each limb, and then uniformly sample points around them.
    \item[$\bullet$] \textbf{GaussianGAN.} We propose an end-to-end pipeline that accepts rendered Gaussian splatting features from a refined point cloud initialisation and novel-view segmentation maps. Then, our generator synthesises photorealistic images with real-time rendering speed. We enforce high-frequency details through a perceptual loss, which has been proven to result in sharper images with high contrast.
    \item[$\bullet$] \textbf{Application.} As far as we know, this is the first neural rendering sign language production system that synthesises photorealistic signers and generalises to unseen sign sequences.
\end{itemize}

\begin{figure*}[t!]
    \centering
    \begin{overpic}[width=0.9\textwidth]{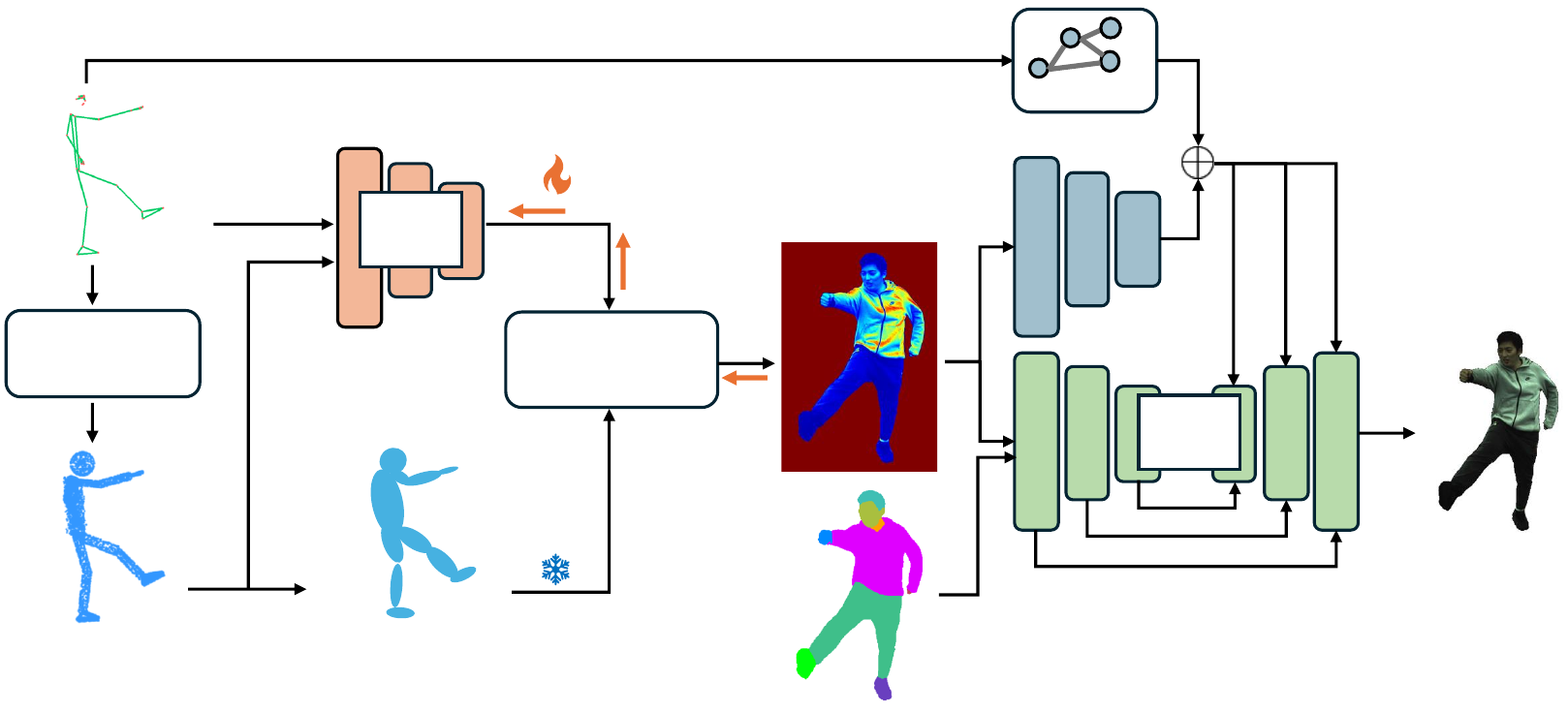}
    

    \put(24.6,29.9){ \Large $\displaystyle \Psi$}

    \put(1.1,23.2){ \footnotesize Densify PCD}
    \put(3.1,21.2){ \footnotesize (Alg.~\ref{alg:densify_pcd})}

    \put(33.6,23.2){ \footnotesize Diff Gaussian}
    \put(34,21.2){ \footnotesize Rasterization}

    \put(34,27.1){ \normalsize $\displaystyle \Delta \mathcal{G}$}
    \put(35.4,15.4){ \normalsize $\displaystyle \mathcal{G}$}

    \put(66.5,39.0){ \footnotesize GNN}

    \put(74.5,17.3){ \normalsize $\displaystyle \mathcal{F}$}

    \put(89,19.0){ \footnotesize $\displaystyle \mathbf{I}_{\text{pred}}$}

    \put(47,25.0){ \footnotesize $\displaystyle \mathbf{F}$}
    \put(47,10.0){ \footnotesize $\displaystyle \mathbf{S}$}

    \put(0,35.0){ \footnotesize $\displaystyle p$}
    \put(0,14.0){ \footnotesize $\displaystyle \mathcal{P}$}

    \put(20.6,37.9){ \footnotesize PCD def. (Sec.~\ref{subsec:def})}
    \put(20.6,2.9){ \footnotesize PCD init. (Sec.~\ref{subsec:pcd})}
    \put(68.3,2.9){ \footnotesize Generator (Sec.~\ref{subsec:gen})}
    
  \end{overpic}
    \caption{\textbf{Overview of our proposed GaussianGAN.} Starting from a 3D skeleton $p$, representing a human in a certain posture. We create a point cloud based on the limbs following our proposed densification strategy as presented in Alg.~\ref{alg:densify_pcd}. The initial Gaussian parameters $\mathcal{G}$ are formed from this point cloud. A pose-guided deformation module consisting of an MLP network regresses a residual $\Delta \mathcal{G}$ of the parameters to better fit the human body to the posture $p$. The Gaussian parameters are then rendered (splatted) onto the image plane of the novel view, together with semantic segmentation maps. These are then passed to a UNet generator to produce photorealistic images.}
    \label{fig:main}
\end{figure*}

The rest of the paper is presented as follows: Sec.~\ref{sec:soa} gives an overview of recent work on neural radiance fields applied to animatable avatars. Sec.~\ref{sec:method} presents our proposed GaussianGAN and justifies the choice of individual components. Sec.~\ref{sec:exp} compares our results against the state of the art. Finally, Sec.~\ref{sec:conc} concludes this paper.

\section{Related Work} \label{sec:soa}
Reconstructing an object, a person or an animal from a 3D scene is a long-standing problem in computer graphics and computer vision~\cite{Yunus_CGF_2024}. Depending on the application and available hardware, the 3D representation of the object of interest can take various forms: Point clouds~\cite{le20bmvc}, voxels~\cite{Kar_2017_neurips}, meshes~\cite{liu2019softras,jimaging7060096}, depth~\cite{Cao2022FWD} and neural fields~\cite{shahariar2023beyondpixels,gao2023nerfneuralradiancefield}. Neural fields have recently attracted the attention of the community due to their ability to represent a scene (\textit{i.e.} geometry and appearance) as a continuous function, typically parameterised by neural networks. The most common paradigms are Neural Radiance Fields (NeRFs)~\cite{mildenhall2020nerf} and 3D Gaussian Splatting (GS)~\cite{kerbl3Dgaussians}.

\subsection{Neural radiance fields} Neural radiance fields represent a scene by a continuous function, often modelled as neural networks, and have been successfully used in applications such as Autonomous Driving~\cite{Zhou_2024_CVPR} or 3D Object Detection~\cite{3dgsdet}. We categorise state-of-the-art methods into implicit and explicit neural radiance fields. Implicit methods are defined by NeRFs~\cite{mildenhall2020nerf}, which encode a scene as a volumetric field defined by its opacity and colour using a multi-layer perceptron (MLP). Through training, the MLP learns to assign opacity and colour to each 3D point in the space of the scene along with the camera orientation using a ray casting technique~\cite{Kato2020DifferentiableRA}. More recently, 3D Gaussian splatting has been proposed as an alternative to NeRFs, which models a scene through an explicit representation modelled by Gaussains and rendered by rasterisation~\cite{Kato2020DifferentiableRA}.

\subsection{Neural Articulated Radiance Field} Neural Articulated Radiance Field or animatable avatars~\cite{Peng_2021_CVPR} learns to render a posed human by learning its shape and appearance through differentiable rendering~\cite{Kato2020DifferentiableRA}. Most methods use a human mesh to drive the rendering, which can be either person-specific~\cite{jiang2022hifecap} or a human template (usually using SMPL)~\cite{Yu_2023_CVPR,hu2023gauhuman,weng_humannerf_2022_cvpr}. Thanks to its out-of-the-box skinning feature~\cite{James_TOG_2005} SMPL is often used for this problem. NeRFs and 3DGS assume a resting human pose (T-pose), often referred to as the canonical pose. In the case of NeRFs~\cite{mildenhall2020nerf}, the idea is to map each 3D point in the observation space to the canonical space and then query the implicit function to estimate the colour and opacity. Gaussian splatting, on the other hand, thanks to its explicit representation, uses a forward flow to map the points in the canonical space to the observation space.

We categorise the Gaussian splatting methods into UV maps, person-specific and SMPL. UV map methods~\cite{Pang_2024_CVPR,hu2024gaussianavatar} use a 2D lookup map for each vertex of the SMPL mesh to learn the Gaussian parameters. Such a method provides flexibility in transferring textures from challenging poses. Person-specific methods~\cite{li2024animatable} build the Gaussian splatting parameters directly from scans of a person (usually from a near-resting pose).
Finally, SMPL-based methods~\cite{qian20233dgsavatar,hu2023gauhuman,Moreau_2024_CVPR} use the human template mesh provided by SMPL~\cite{James_TOG_2005} to initialise the Gaussian splats, providing a standard representation that can be generalised across individuals. Some methods take into account the correction of the SMPL parameters to better fit the Gaussian in canonical space~\cite{qian20233dgsavatar,hu2023gauhuman}. The regularisation of the skinning weight can also lead to a better deformation~\cite{Moreau_2024_CVPR,kocabas2024hugs}. In addition, some methods~\cite{shao2024splattingavatar,svitov2024haha,wen2024gomavatar,jena2023splatarmor} rely on the SMPL mesh to learn the parameters of the Gaussian relative to the face position.

\section{Method} \label{sec:method}
In this section, we present our GaussianGAN model. First, we give an overview of Gaussian splatting, which is used both for the novel-view segmentation and for the representation of features of the person in the target body pose. Then, we present our algorithm for point-cloud densification. Finally, we describe our two-stage pipeline (pose-guided deformation and generator) for synthesising photorealistic human avatars. An overview of our proposed GaussianGAN model can be found in Fig~\ref{fig:main}.

\subsection{Gaussian Splatting} \label{subsec:gso} Gaussian Splatting is an explicit representation that models a scene using 3D Gaussians~\cite{kerbl3Dgaussians}. The Gaussian is characterised by a covariance matrix $\Sigma$ and a centre (mean value) point $\mu \in \mathbb{R}^3$ as:
\begin{equation}
    G(x) = \exp \Big \{ -\frac{1}{2}(x-\mu)^{\top} \Sigma^{-1}(x-\mu) \Big \}.
\end{equation}

In addition, the Gaussian is parameterised with a $3$D-rotation $\mathbf{q} \in \mathbb{R}^4$ (represented as a quaternion), a scaling factor $\mathbf{s} \in \mathbb{R}^3$, an opacity $\mathbf{o} \in \mathbb{R}$, and a view-dependent colour $\mathbf{c} \in \mathbb{R}^3$, which is defined with spherical harmonics (SH). To enable stable optimisation during training, the covariance matrix $\Sigma$ is defined as positive semi-definite:
\begin{equation}
    \Sigma = \mathbf{R}\mathbf{S}\mathbf{S}^{\top}\mathbf{R}^{\top},
\end{equation}
where $\mathbf{S} = \texttt{diag}([s_x, s_y, s_z])$, and the rotation matrix $\mathbf{R} \in SO(3)$ transformed from $\mathbf{q}$. 
These Gaussians are then rendered into the scene using Elliptical Weighted Average (EWA) splatting~\cite{Ren_2002_OSE}. The covariance matrix is projected onto the camera coordinates given a viewing transformation $\mathbf{W} \in \mathbb{R}^{3 \times 3}$ (also known as the camera pose) such that:

\begin{equation}
    \Sigma_{2D} = \mathbf{J}\mathbf{W}\Sigma\mathbf{W}^{\top}\mathbf{J}^{\top},
\end{equation}
where $\mathbf{J} \in \mathbb{R}^{2 \times 3}$ is the Jacobian of the projective transformation and $\Sigma_{2D}$ is the projected (Splatted) $2$D-Gaussian. Finally, the colour $\mathbf{C}$ is evaluated at pixel position $\mathbf{x} \in \Omega$ in the image plane of size $\Omega = \{1, \dots, W\} \times \{1, \dots, H \}$, where $W$ (resp. $H$) is the image width (resp. height). The colour is calculated by point-based cumulative volumetric rendering~\cite{Kato2020DifferentiableRA} using $\mathcal{N}$ ordered points that overlap at this pixel position such that:
\begin{equation}
    \mathbf{C}(\mathbf{x}) = \sum_{k \in \mathcal{N}} \mathbf{c}_k \alpha_k \left[ \prod_{i=1}^{k-1}(1-\alpha_i) \right],
\end{equation}
where $\alpha_k$ is obtained by $\Sigma_{2D}$ multiplied by each Gaussian opacity $\mathbf{o}_k$.

\subsection{Point-cloud initialisation} \label{subsec:pcd} Gaussian splatting processes point clouds from a dense multi-view environment using Structure from Motion (SfM) ~\cite{Johannes_2016_CVPR} or from a monocular or sparse environment using a prior such as a human template like the Skinned Multi-Person Linear Model (SMPL) ~\cite{Loper_2015_TOG}. However, in a sparse environment, SfM tools such as COLMAP fail. SMPL estimation is also prone to errors, especially with complex hand movement. The reason for this failure is mainly due to depth ambiguity~\cite{Belhumeur_1997_CVPR,uy-scade-cvpr23}, where many 3D objects represent the same 2D shape.

To overcome these limitations, we propose a novel point cloud densification strategy based on the connection between the limbs of the human body. Good point cloud initialisation in Gaussian splatting contributes to the overall rendering performance~\cite{kerbl3Dgaussians}. This is because by providing accurate scene geometry, learning starts with a solid baseline and so focuses on the appearance of the scene with small changes in the Gaussian position. To construct our point cloud using a $3$D human skeleton, we build a 3D cylinder around each limb and sample points from the cylinder surface. Such a strategy allows us to create a good initialisation of the Gaussians around these points. It is worth noting that we create the point cloud directly from the 3D skeleton and thus we do not require a skinning mechanism.

Our proposed point cloud densification strategy follows the shape of the human body. We start from a 3D skeleton (Fig.~\ref{fig:densify_pcd}(a)) and create a point cloud that follows the limbs of the body (Fig.~\ref{fig:densify_pcd}(b)). Specifically, we select points around each limb from a template of human skeleton keypoints (\textit{e.g.} Mediapipe~\cite{mediapipe_2019_arxiv}) to create the point cloud (Alg.~\ref{alg:densify_pcd}). First, we build a cylinder around each limb using the dictionary$\mathcal{J}$ of joint connections from the template skeleton. The radius of the cylinder is empirically set to $0.1$ times the norm of the distance between the two points $p_i, p_j$ that define the limb. We take $200$ points uniformly around the cylinder. Note that for the facial region (resp. torso), we instead build a sphere (resp. cuboid) whose radius is the norm of the longest distance between the 3D points of the facial landmarks (resp. torso), and then sample $500$ points. The result of this procedure is a point cloud $\mathcal{P} = \{ \mu_i \}_{i=1}^N$ of size $|\mathcal{P}| = N$, which is used to form the initial centres of our Gaussian. Due to the simplicity of our approach, we can model any posture.

\subsection{Pose-guided deformation} \label{subsec:def} Given a $3$D-skeleton $p$ of a human posture and the corresponding point cloud $\mathcal{P}$ resulting from our proposed Alg.~\ref{alg:densify_pcd}, we initialise a set of Gaussian splatting nodes with the default parameters \mbox{$\mathcal{G} = \{ \mathcal{G}_k = (\mu_k, \mathbf{q}_k, \mathbf{s}_k, \mathbf{o}_k, \mathbf{c}_k, \textcolor{blue}{ \mathbf{f}_k})) \}_{k=1}^{|\mathcal{P|}}$}. Note that, in contrast to SMPL-based models~\cite{Loper_2015_TOG}, the number of point in the cloud $N$ is fixed; the reason is the absence of a skinning mechanism. Therefore, we decide to freeze the learning of $\mathcal{G}$ and learn a deformation $\Psi$ instead.

From the estimated skeleton $p$ we can accurately obtain an initial point cloud representing the underlying mesh. Therefore, it is sufficient to learn a residual deformation field $\Delta \mathcal{G}$ instead of the actual Gaussian $\mathcal{G}$ (Fig.~\ref{fig:main}).
Therefore, we model the deformation by a mapping $\Psi$ that regresses the deformation residual from the initial parameters extracted from the point cloud $\mathcal{P}$. For each of the Gaussians $\mathcal{G}_k \in \mathcal{G}$ we estimate the residuals as: \mbox{$\Psi(\Delta \mathcal{G}| p, \mathcal{P}) = (\delta \mu, \delta \mathbf{q}, \delta \mathbf{s}, \delta \mathbf{o}, \delta \mathbf{c}, \delta \textcolor{blue}{ \mathbf{f}})$}.

Now we combine the estimated deformations for each Gaussian $\mathcal{G}_k \in \mathcal{G}$ as: $\hat{\mu}_k = \mu_k + \delta \hat{\mu}_k$, $\hat{\mathbf{q}}_k = \mathbf{q}_k + \delta \hat{\mathbf{q}}_k$, $\hat{\mathbf{s}}_k = \mathbf{s}_k + \delta \hat{\mathbf{s}}_k$,
$\hat{\mathbf{o}}_k = \mathbf{o}_k + \delta \hat{\mathbf{o}}_k$,
$\hat{\mathbf{c}}_k = \mathbf{c}_k + \delta \hat{\mathbf{c}}_k$, and $\textcolor{blue}{\hat{\mathbf{f}}_k} = \textcolor{blue}{\mathbf{f}_k} + \delta \textcolor{blue}{\hat{\mathbf{f}}_k}$. 

\subsection{Novel-view segmentation module} \label{subsec:seg} CNN encoder-decoder generators such as UNet require more modalities to output sharper results \textit{e.g.} DensePose~\cite{Neverova_2018_ECCV}, Semantic Segmentation~\cite{song2019unsupervised}, and SMPL~\cite{lwb2019}. Our generator uses segmentation as a modality as it provides accurate masking and helps the network to learn a better representation per semantic region. Importantly, our approach can be adapted to any deformable object, such as animals~\cite{Zuffi_CVPR_2018}. 

It is important to note that the CNN generator can accept different representations of the segmentation maps depending on the task. For animatable avatars, for example, the body posture moves with many more degrees of freedom and therefore a complete segmentation map of the body is required. However, in sign language production, 2D skeletons have been shown to be sufficient to capture the manual and non manuals~\cite{Saunders_2022_CVPR}. This is because only the hands and face are essential. Therefore, in SLP we use the segmentation of the hands instead.

\begin{figure}[t!]
    \centering
    \begin{overpic}[width=1.\linewidth]{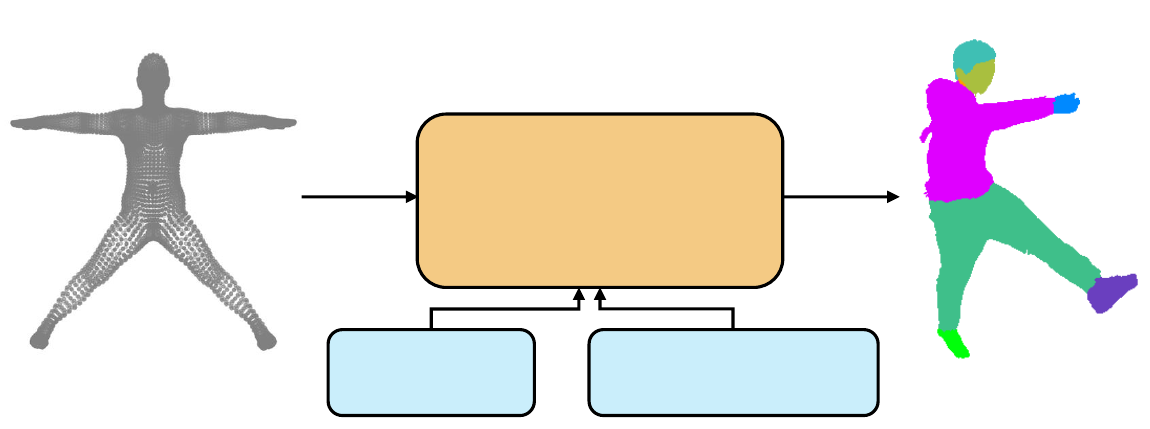}

    \put(38.6,22.9){ \footnotesize 3D-GS novel-view }
    \put(36.6,19.1){ \footnotesize Segmentation module}
    \put(44.6,14.9){ \footnotesize (Sec.~\ref{subsec:seg})}

    \put(27.9,4.1){ \footnotesize Body pose $\theta$}
    \put(54.8,5.5){ \footnotesize Observation}
    \put(51.1,2.4){ \footnotesize space calibration}

    \put(1.9,2.1){ \footnotesize Canonical space}
    \put(76.9,2.1){ \footnotesize Observation space}
    
  \end{overpic}
    
    \caption{\textbf{Novel-view segmentation module.} We render novel-view semantic segmentation using Gaussian splatting~\cite{kerbl3Dgaussians}. In particular, we use an SMPL-based approach to render the segmentation on the observation space given a target body pose $\theta$.}
    \label{fig:nvs_seg}
\end{figure}

In the following, we describe an SMPL-based module to represent the semantic segmentation of a person from different postures (Fig.~\ref{fig:nvs_seg}) in the novel view.~\footnote{Note that the Novel-view segmentation module does not depend on a specific paradigm \textit{e.g.} SMPL. Here, SMPL can be a good choice, since the problem of depth ambiguity only occurs with fine-grained details such as the interlacing of hands.}
In particular, given a canonical SMPL model in resting $T$-pose, we initialise the segmentation point cloud $\mathcal{P}^s$ from the vertices of the SMPL of size $|\mathcal{P}^s| = 6890$. For each vertex in the canonical space $\mathbf{x}_c \in \mathcal{P}^s$ we assign the corresponding Gaussian parameters: $\mathcal{G}^s = (\mathbf{x}_c, \mathbf{q}_c, \mathbf{s}_c, \mathbf{s}_c)$, where $\mathbf{s}_c$ is the corresponding semantic segmentation class. For the target body pose $\theta$, we now use Linear Blend Skinning (LBS)~\cite{Loper_2015_TOG} to deform the canonical point $\mathbf{x}_c$ into the observation space:

\begin{align*}
    \mathbf{x}_o =\text{LBS}(\mathbf{x}_c, \{\mathbf{\Delta B}_b\}^B_{b=1}, \omega_{\mathbf{x_c}})=(\sum_{b=1}^{B}\omega_{\mathbf{x_c}}^{b}\mathbf{\Delta B}_b)\mathbf{x}_c,
    \label{eq_LBS}
\end{align*}
where $\{\mathbf{\Delta B}_b\}^B_{b=1}$ represents the rigid bone transformations of the SMPL template, and $\omega_{\mathbf{x}_c}^b$ denotes the skinning weight of the point $\mathbf{x}_c$ associated with bone $\mathbf{B}_b$.

Finally, using the calibration from the observation space, we render the semantic segmentation from $\mathcal{G}^s$ using differentiable Gaussian rasterisation as:

\begin{equation}
    \mathbf{S}(\mathbf{x}) = \sum_{k \in \mathcal{N}} \mathbf{s}_k \alpha_k \left[ \prod_{i=1}^{k-1}(1-\alpha_i) \right].
\end{equation}
To train our network, we use a cross entropy loss between the rendered segmentation and the ground-truth. 

\subsection{Generator} \label{subsec:gen}  The calibration of the observation space is represented by its intrinsic $\mathbf{K}$, extrinsic $\mathbf{R}$ and translation $\mathbf{t}$. Using the Gaussian differential rasterisation as defined in~\cite{kerbl3Dgaussians}, we render the features $\textcolor{blue}{\hat{\mathbf{f}}} \in \mathcal{G}$ to the image plane of the observation space as: \mbox{$\mathbf{F} = \texttt{rasterise}(\hat{\mathcal{G}}, \mathbf{K}, [\mathbf{R}, \mathbf{t}])$}. Similarly, we render the trained Gaussian splatting segmentation model as \mbox{$\mathbf{S} = \texttt{rasterise}(\mathcal{G}^s, \mathbf{K}, [\mathbf{R}, \mathbf{t})]$}. To render the person in motion from a novel-view camera, we combine $\mathbf{F}$ and $\mathbf{S}$ through a UNet~\cite{Xiaomeng_2018_TMI} encoder-decoder model. Since such a model retains the spatial information from the conditional input, the model would only focus on learning the pixel distribution within the novel view segmentation $\mathbf{S}$.

Let $\mathcal{E}$ represents the encoding module of the UNet model labelled $\mathcal{F}$, we use two separate encoders, namely one, that combines $\mathbf{F}$ and $\mathbf{S}$ and the projection of the $3$D skeleton $p$ onto the observation space $\mathbf{p}_{2\text{D}}$ by concatenation $\mathcal{E}_c(\mathbf{F} \oplus \mathbf{S} \oplus \mathbf{p}_{2\text{D}})$. We further encode $\mathbf{F}$ to obtain features in different resolutions using a dedicated encoder $\mathcal{E}_s$. In addition, we encode the $3$D-skeleton of the human pose, which serves as a global representation for the novel-view synthesis. In particular, we create a $6$D-rotation~\cite{Zhou_2019_CVPR} from each joint of the pose $p$, such that: \mbox{$\Omega = \{ \omega_1, \dots, \omega_J \}$}, where $\omega_k \in \mathbb{R}^6$. We then use a Graph Neural Network (GNN) as defined in~\cite{Su_2022_ECCV} to encode $\Omega$, and the feature representation of $p$ is therefore given as: \mbox{$\mathbf{g} = \text{GNN}(\Omega)$}. The combination of $\mathbf{g}$ and the multilayer features from $\mathcal{E}_s$ now serves as the style for our UNet model (Fig.~\ref{fig:main}). The novel-view image $\mathbf{I}_{\text{pred}}$ is then obtained by the UNet decoder (Alg.~\ref{alg:gaussianGAN_train}).~\footnote{It is worth noting that the pipeline presented is end-to-end trainable, and the errors coming from the $\ell_2$ reconstruction loss from the encode-decoder model $\mathcal{F}$ help to learn better Gaussian feature maps $\mathbf{F}$.}

\begin{algorithm}[H]
\footnotesize
\caption{Training of GaussianGAN}
\label{alg:gaussianGAN_train}
  \begin{algorithmic}
    \State \textbf{Inputs:} $\mathcal{F}$: a UNet encoder-decoder generator. $p = \{ p_i \}_{i=1}^J$: $3$D human skeleton of $J$ joints. $\mathbf{I}$: target image. $\mathbf{K}$: intrinsic camera.\\ $\mathbf{R}$: extrinsic camera. $\mathbf{t}$: translation matrix.
     
    \State $\mathcal{P} \gets \texttt{densify}(p)$ \Comment{Initial point-cloud (Alg.~\ref{alg:densify_pcd}).}
    
    \State $(\delta \mu, \delta \mathbf{q}, \delta  \mathbf{s}, \delta  \mathbf{o}, \delta  \mathbf{c}, \delta \textcolor{blue}{ \mathbf{f}}) \gets \Psi(\Delta \mathcal{G}| p, \mathcal{P})$
    
    \State $\mathbf{F} \gets \texttt{rasterise}(\hat{\mathcal{G}}, \mathbf{K}, [\mathbf{R}, \mathbf{t}])$
    
    \State $\mathcal{L}_{\Psi} \gets (1-\lambda) \mathcal{L}_1 + \lambda \mathcal{L}_{\text{D-SSIM}}$ \Comment{\textcolor{cyan}{\text{Step $I$ optimisation.}}}

    \State $\mathbf{S} \gets \texttt{rasterise}(\mathcal{G}^s, \mathbf{K}, [\mathbf{R}, \mathbf{t})]$

    \State $\mathbf{I}_{\text{pred}} \gets \mathcal{F}(\mathbf{F}, \mathbf{S})$

    \State $\mathcal{L} \gets \mathcal{L}_{\text{GAN}} + \mathcal{L}_{\text{perc}} + \mathcal{L}_{\text{feat}}$ \Comment{\textcolor{cyan}{\text{Step $II$ optimisation.}}}
        
    \end{algorithmic}
\end{algorithm}

\textbf{Training.}
To train the deformation mapping $\Psi$, we use an $\ell_1$ reconstruction loss to account for low frequencies and SSIM~\cite{Wang_2004_TIP} to account for high frequencies as:

\begin{equation}
    \mathcal{L}_{\Psi} = (1-\lambda) \mathcal{L}_1 + \lambda \mathcal{L}_{\text{D-SSIM}},
\end{equation} where $\lambda$ is a controlling factor as defined in~\cite{kerbl3Dgaussians}. Note that $\mathcal{L}_{\text{D-SSIM}}$ serves as a regulariser for the feature maps $\mathbf{F}$.

The second step of training consists of a reconstruction loss $\mathcal{L}_{\text{perc}}$ using the perceptual network~\cite{Johnson_2016_ECCV} and the GAN loss~\cite{goodfellow2014generative} to ensure that the generated images look photorealistic. Additionally, we add a multiscale discriminator loss~\cite{Rombach_2022_CVPR} $\mathcal{L}_{\text{feat}}$ to force high-frequency details at different image scales. This has proven to be an effective way to overcome the blurring of the rendered images that the NeRF and GS models suffer from.

\begin{table*}[htb!]
    \centering
    \resizebox{\textwidth}{!}{
    \begin{tabular}{lccc|ccc|ccc|ccc|ccc|ccc}
    \toprule
      \multirow{2}{*}{\textbf{Method}}  & \multicolumn{3}{c|}{\textbf{377}} & \multicolumn{3}{c|}{\textbf{386}} & \multicolumn{3}{c|}{\textbf{387}} & \multicolumn{3}{c|}{\textbf{392}} & \multicolumn{3}{c|}{\textbf{393}} & \multicolumn{3}{c}{\textbf{394}}   \\
    \cline{2-19}
        & \textbf{PSNR}  & \textbf{SSIM}  & \textbf{LPIPS}  & \textbf{PSNR}  & \textbf{SSIM}  & \textbf{LPIPS}  & \textbf{PSNR}  & \textbf{SSIM}  & \textbf{LPIPS}  & \textbf{PSNR}  & \textbf{SSIM}  & \textbf{LPIPS}  & \textbf{PSNR}  & \textbf{SSIM}  & \textbf{LPIPS}  & \textbf{PSNR}  & \textbf{SSIM}  & \textbf{LPIPS}   \\
    \midrule

NeuralBody~\cite{peng2021neural} &                      29.11 & 0.97 & 0.04 &                      30.54 & 0.97 &  0.05 &                      27.00 &  0.95 &  0.06 &                      30.10 & 0.96 & 0.05 & 28.61 & 0.96 &  0.06 &  29.10 & 0.96 &  0.05 \\

HumanNerf~\cite{weng_humannerf_2022_cvpr}  &  30.41 & 0.97 &  0.02 & 33.20 &  0.98 & 0.03 & 28.18 & 0.96 & 0.04 &  31.04 &  0.97 &  0.03 &                      28.31 & 0.96 & 0.04 & 30.31 & 0.96 &  0.03  \\ 
ARAH~\cite{Shaofei_2022_ECCV} & 30.85 & 0.98 & 0.03  & 33.50 & 0.98 & 0.03 & 28.49 & 0.97 & 0.04 & 32.02 & 0.97 & 0.04 & 28.77 & 0.97 & 0.04 & 29.46 & 0.96 & 0.04 \\ 
MonoHuman~\cite{Yu_2023_CVPR}  & 30.77 &  0.98 &  0.02 &  32.97 & 0.97 & 0.03 &  27.93 & 0.96 &  0.03 & 31.24 &  0.97 &  0.03 &  28.46 & 0.96 &  0.03 &                      28.94 & 0.96 & 0.04  \\ \hdashline
HUGS~\cite{kocabas2024hugs}   &  30.80 &  0.98 &  0.02 &  34.11 &  0.98 &  0.02 &  29.29 &  0.97 &  0.03 &  31.36 &  0.97 &  0.03 &  29.80 &  0.97 &  0.03 &  30.54 &  0.97 &  0.03  \\
GauHuman~\cite{hu2023gauhuman}   &  32.25 &  0.98 &  0.02 &  33.73 &  0.97 &  0.03 &  28.18 &  0.96 &  0.04 &  32.27 &  0.97 &  0.03 &  30.24 &  0.96 &  0.04 &  31.46 &  0.96 &  0.03  \\
3DGS-Avatar~\cite{qian20233dgsavatar} & 30.64 & 0.98 & 0.02 & 33.63 & 0.98 & 0.03 & 28.33 & 0.96 & 0.03 & 31.66 & 0.97 & 0.03 & 28.88 & 0.96 & 0.04 & 30.54 & 0.97 & 0.03 \\ 

\midrule
Ours   &  \textbf{34.16} &  \textbf{0.99} &  \textbf{0.02} &  \textbf{36.45} &  \textbf{0.98} &  \textbf{0.02} &  \textbf{29.99} &  \textbf{0.97} &  \textbf{0.03} &  \textbf{33.23} &  \textbf{0.98} &  \textbf{0.02} &  \textbf{31.50} &  \textbf{0.97} &  \textbf{0.03} &  \textbf{32.26} &  \textbf{0.97} &  \textbf{0.03}  \\

    \bottomrule
    \end{tabular}  
    }
    \caption{ \textbf{Quantitative comparison on ZJU-MoCap Dataset.} We evaluate the performance on novel pose synthesis. We quantify the performance of the model using the metrics PSNR, SSIM and LPIPS. Our model achieves the best performance in all scenes and metrics.}
    \label{tab:zju}
\end{table*}

\section{Experiments} \label{sec:exp}
In this section, we present the experimental evaluation of our proposed approach compared to current state-of-the-art methods, achieving superior photorealistic quality results at a real-time rendering speed of 79 FPS. We then ablate each component of our proposed model to highlight its importance to the overall rendering quality.

Our GaussianGAN is trained on a single RTX3090 graphics card and with the Adam~\cite{Kingma_2015_ICLR} optimiser with a learning rate of $2e^{-5}$ for the GAN generator. We train the Gaussian splatting parameters for 10k iterations in step I (Alg.~\ref{alg:gaussianGAN_train}) as they converge faster. The entire training of our pipeline takes about ten hours.

\subsection{Datasets and Metrics}
\textbf{ZJU-MoCap Dataset.} The dataset is a widely used benchmark for controllable human avatars in the community due to its challenging camera views, posture and texture~\cite{peng2021neural}. Following~\cite{weng_humannerf_2022_cvpr}, we choose six human subjects (377, 386, 387, 392, 393, 394). Of the total number of cameras available, we use one camera for training and the remaining 22 for testing.

\textbf{THuman4 Dataset.} We follow the same evaluation protocol as suggested by PoseVocab~\cite{li2023posevocab}. In particular, we used the sequence “subject00” recorded with 24 cameras. To evaluate the novel pose synthesis, each method is trained with 23 cameras and tested with the remaining camera. We use the first 2000 frames for training and evaluate the remaining 500 frames.

\textbf{Evaluation Metrics.} To quantitatively evaluate the performance of each model, we use the pixel-based metrics Peak Signal-to-Noise Ratio (PSNR) and Structural Similarity (SSIM)~\cite{Wang_2004_TIP} and the feature-based metric Learned Perceptual Image Patch Similar (LPIPS)~\cite{zhang2018perceptual}.

\subsection{Comparison with the state-of-the-art methods}
\textbf{Baselines.} We compare our proposed GaussianGAN against neural rendering methods. Note that there are no common datasets used by all methods in the literature. Therefore, we report two different sets of methods per dataset. For ZJU-Mocap dataset, we compared against both NeRF methods~\cite{peng2021neural,weng_humannerf_2022_cvpr,Shaofei_2022_ECCV,Yu_2023_CVPR} and Gaussian splatting methods~\cite{kocabas2024hugs,hu2023gauhuman,qian20233dgsavatar}. For the THuman4 dataset, we compare against the following Nerf methods~\cite{Li_2023_SIGGRAPH,Zheng_2022_CVPR,Li_2022_ECCV,Peng_2021_ICCV,Shaofei_2022_ECCV,Liu_2024_CVPR} and and HuGS~\cite{Moreau_2024_CVPR}, which uses Gaussian splatting.

\textbf{Methods for Comparison.} For qualitative evaluation, we compare our method
against (1) ARAH~\cite{Shaofei_2022_ECCV} which the best performing NeRF-based method; and (2) GauHuman~\cite{hu2023gauhuman}, which is a Gaussian splatting approach.

\begin{figure}[h!]
    \centering
    \includegraphics[width=1.\linewidth]{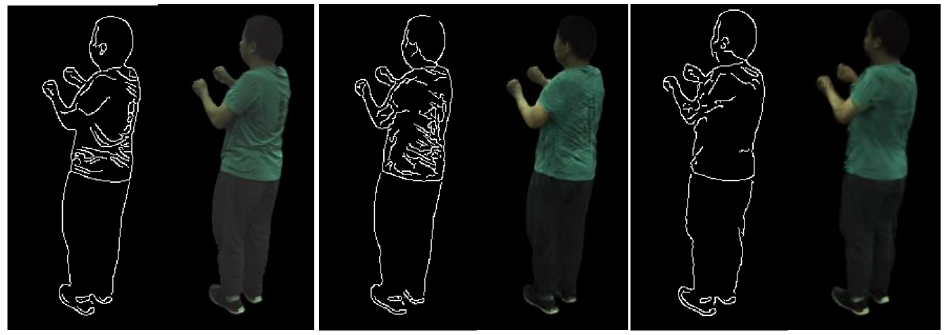}
    \caption{\textbf{High-frequency preservation.} Our proposed model preserves the high frequency details of the body in the novel-view and novel-pose compared to neural radiance fields approaches. We highlight this through canny edge detection. KEY -- (left) GT; (middle) ours; (right) GauHuman~\cite{hu2023gauhuman}.}
    \label{fig:soa_histo_comp}
\end{figure}

\textbf{Comparison Results.} Our proposed method runs in real-time at 79 FPS and is comparable to recent Gaussian splatting methods that run at rendering speed of 30~\cite{Pang_2024_CVPR} 35~\cite{hu2024gaussianavatar}, 43~\cite{wen2024gomavatar}, 50~\cite{qian20233dgsavatar}, 60~\cite{kocabas2024hugs}, and 80~\cite{Moreau_2024_CVPR} FPS.

\begin{figure*}[h!]
    \centering
    \begin{overpic}[width=1.\linewidth]{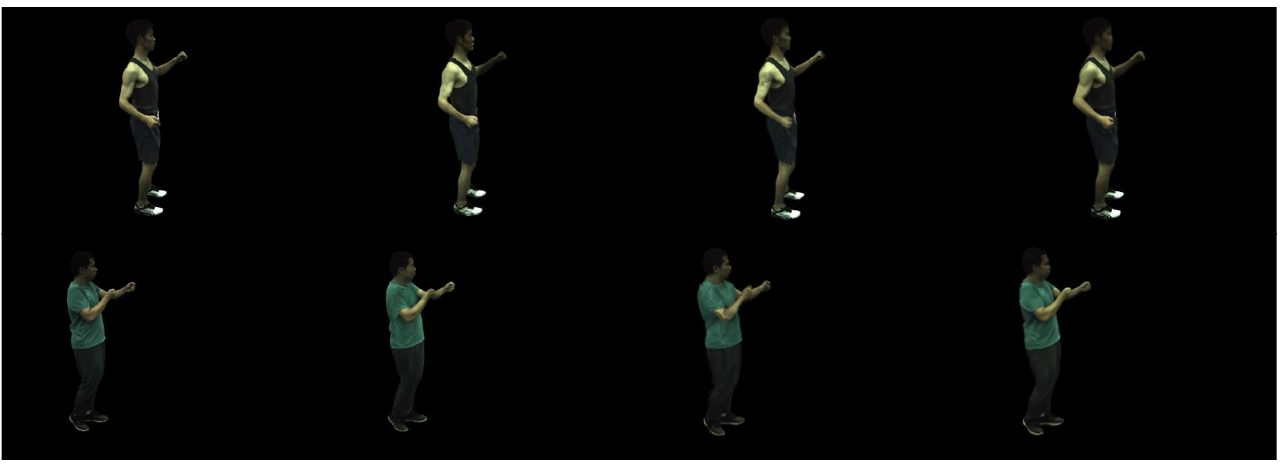}
    \put(10,-1.2){ \footnotesize GT}
    \put(34,-1.2){ \footnotesize Ours}
    \put(56,-1.2){ \footnotesize GauHuman~\cite{hu2023gauhuman}}
    \put(83,-1.2){ \footnotesize ARAH~\cite{Shaofei_2022_ECCV}}
    
  \end{overpic}
    \caption{\textbf{Qualitative comparison} of our proposed GaussianGAN against a Gaussian splatting method (GauHuman~\cite{hu2023gauhuman}) and a NeRF approach (ARAH~\cite{Shaofei_2022_ECCV}). The results show novel-view test examples on the ZJU-MoCap dataset~\cite{Peng_2021_CVPR}.}
    \label{fig:soa_comp}
\end{figure*}

\begin{figure*}[h!]
    \centering
    \includegraphics[width=1.\linewidth]{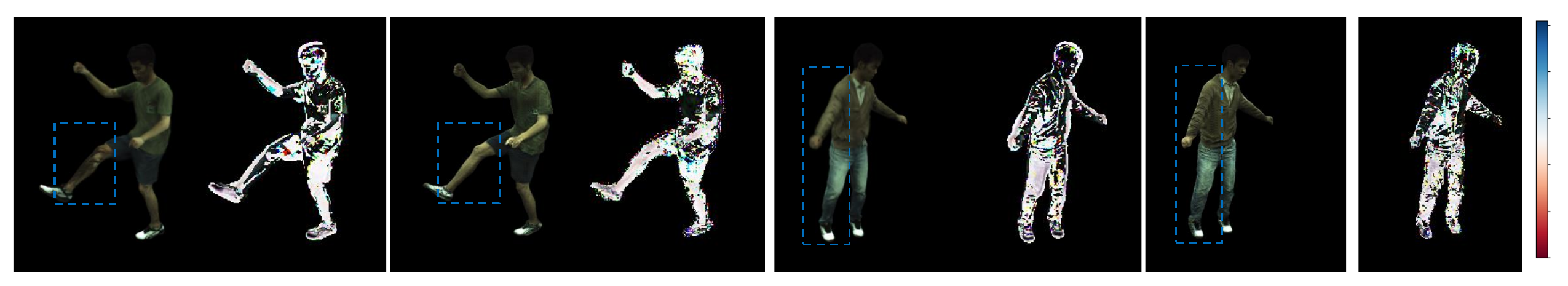}
    \caption{\textbf{GaussianGAN vs. neural rendering (only)}. We highlight some of the key features that our proposed model takes into account, namely blurriness and details preservation. KEY -- (left) GauHuman~\cite{hu2023gauhuman}; (right) ours.}
    \label{fig:soa_heatmap_cmp}
\end{figure*}

Tab.~\ref{tab:zju} shows the results of our method compared to the above methods on the ZJU-MoCap dataset. We find that our method achieves the best results in synthesising novel views in all metrics. We also report competitive results on the THuman4 dataset when synthesising novel-pose (Tab.~\ref{tab:thuman_quantitative}). We argue that the success of our method compared to neural rendering approaches (NeRFs and GS) is due to the fact that the UNet generator, together with its losses, both favours high frequencies and helps to preserve the details of the body in the novel pose/view. Fig.~\ref{fig:soa_heatmap_cmp} shows an example that emphasises this claim. We can see that our method produces sharper results and preserves the details of the clothing in the novel view. Fig.~\ref{fig:soa_histo_comp} shows an example of applying Canny Edges to the synthesised images and the ground-truth. We can clearly see that our approach has more edges, which confirms the claim that more high-frequency details are generated. Finally, Fig.~\ref{fig:soa_comp} shows more qualitative results. Our model has fewer visible artefacts and a better texture than GauHuman~\cite{hu2023gauhuman} and ARAH~\cite{Shaofei_2022_ECCV}. In particular, GauHuman produces visible artefacts and less detail around the hands and arms. While ARAH delivers blurred results.

\subsection{Ablation Study} To assess the effectiveness of our proposed pipeline, we conducted the following ablation to highlight the benefit of each of the proposed component. 

\textbf{3D-aware global feature.} We highlight the importance of our 3D-aware global feature by  replacing the GCN module with a linear layer such that: $\mathbf{g} = \mathbf{W} p + b$ such that $\mathbf{W} \in \mathbb{R}^{99 \times 35}, b \in \mathbb{R}^{35}$. We can see that adding the graph convolutional network to encode the pose further improves the synthesis quality from 29.08 to 33.17 PSNR (Tab.~\ref{tab:ablation}).

\begin{figure}[h!]
    \centering
    \includegraphics[width=1.\linewidth]{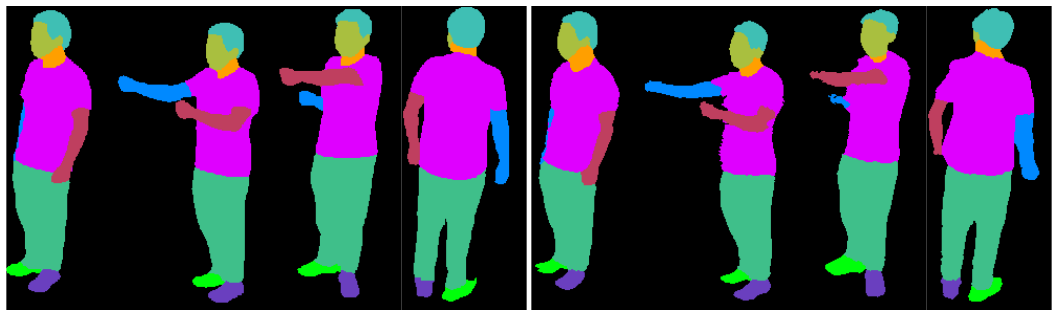}
    \caption{\textbf{Novel-view segmentation.} Qualitative results of our proposed segmentation module (Sec.~\ref{subsec:seg}) on novel-view and novel-pose sequences. Our module can accurately render novel view segmentation. KEY -- (left) GT segmentation; (right) our segmentation results.}
    \label{fig:nvs_seg_smpl_exp}
\end{figure}

\textbf{Novel-view segmentation.} Our proposed segmentation module outputs segmentation maps with high mIoU values (Tab.~\ref{tab:nvs_seg_res}). We have tested our model in novel-view novel-pose segmentation and our model is indeed capable of generating accurate maps (Fig.~\ref{fig:nvs_seg_smpl_exp}). Furthermore, to assess the benefit of using the novel-view segmentation maps $\mathbf{S}$, we also removed them from our GAN generator (Tab.~\ref{tab:ablation}). We can see that $\mathbf{S}$ actually helps the generator to focus on the texture, as the geometry is mostly given by the segmentation.

\textbf{Feature loss.} We highlight using the multiscale discriminator loss~\cite{Rombach_2022_CVPR} with GaussianGAN. Results from Tab.~\ref{tab:ablation} show that the GAN model can produce better results with the loss as translated by the higher PSNR scores.

\begin{figure}
    \centering
    \begin{overpic}[width=.5\textwidth]{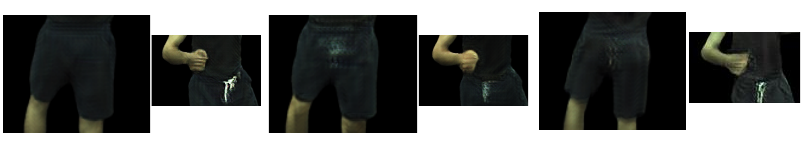}
    \put(10,-2){ \footnotesize Full model}
    \put(42,-2){ \footnotesize w/o GCN}
    \put(77,-2){ \footnotesize w/o $\mathbf{S}$}
    
  \end{overpic}
    \caption{\textbf{Ablation}. We highlight the importance of using different modalities to improve the synthesis quality. KEY -- (right): generator w/o segmentation mask $\mathbf{S}$; (middle): generator using $\mathbf{S}$, but w/o GCN; (left): generator with both $\mathbf{S}$ and GCN.}
    \label{fig:ablation}
\end{figure}

We show the results of our ablation in Fig.~\ref{fig:ablation}. We can see that without the novel-view segmentation $\mathbf{S}$, the generator struggles with the body shape and produces less detail. The model can focus on getting more details when the semantic segmentation is added. However, the real advantage of GCN lies in the overall body orientation to the camera. Since the pose feature $\Omega$ together with the camera calibration provides a global positioning of the body, the network learns to synthesise the clothing with respect to the camera orientation.

\begin{table}[t!]
    \footnotesize
    \centering
    \caption{ \textbf{Quantitative comparison on Thuman4 Dataset.} We evaluate the performance on novel pose synthesis. We quantify the performance of the model using the metrics PSNR, SSIM and LPIPS. Our model achieves the best performance PSNR and the third best in SSIM and PSNR, respectively.}
    
    \label{tab:thuman_quantitative}
    \begin{tabular}{lccc}
    \hline
    \textbf{Method} & \textbf{PSNR} & \textbf{SSIM} & \textbf{LPIPS}  \\
    \hline
    PoseVocab~\cite{Li_2023_SIGGRAPH} & 30.97 & 0.977 & 17 \\
    SLRF~\cite{Zheng_2022_CVPR} & 26.15 & 0.969 & 24 \\
    TAVA~\cite{Li_2022_ECCV}  & 26.61 & 0.968 & 32 \\
    Ani-NeRF~\cite{Peng_2021_ICCV} & 22.53 & 0.964 & 34 \\
    ARAH~\cite{Shaofei_2022_ECCV} & 21.77 & 0.958 & 37  \\ 
    TexVocab~\cite{Liu_2024_CVPR} & 32.09 & 0.983 & \textbf{13.40} \\ \hdashline
    HuGS~\cite{Moreau_2024_CVPR} & 32.49 & \textbf{0.984} & 19 \\
    \midrule
    Ours & \textbf{33.39} & 0.981 & 17.97  \\
    \hline
    \end{tabular}
    
\end{table}

\begin{table}[t!]
    \footnotesize
    \centering
    \caption{\textbf{Ablation.} This table shows the effects of the individual components of our method. We highlight the importance of the high-frequency loss~\cite{Rombach_2022_CVPR} with $\mathcal{L}_{\text{feat}}$ and show the advantage of our global 3D-aware feature using GCN as well as the novel-view segmentation maps $\mathbf{S}$. We performed our ablation on sequence 377 of the ZJU-MoCap dataset~\cite{Peng_2021_CVPR}.}
    \label{tab:ablation}
    \begin{tabular}{lccc}
    \hline
    \textbf{Method} & \textbf{PSNR} & \textbf{SSIM} & \textbf{LPIPS}  \\
    \hline
    w/o $\mathcal{L}_{\text{feat}}$ & 32.53 &  0.98 & 0.02 \\ \hdashline
    w/o $\mathbf{S}$ & 29.08 &  0.96 & 0.03 \\
    w/o GCN & 33.17 &  0.98 & 0.02 \\ \hdashline
    Full & 34.16 & 0.99 & 0.02 \\
    \hline
    \end{tabular}
    
\end{table}

\begin{table}[t!]
    \footnotesize
    \centering
    \caption{\textbf{Novel-view segmentation} We report the mean Intersection over Union (mIoU) on novel-view and novel-pose segmentation with our module (Sec.~\ref{subsec:seg}). We conducted our experiment using sequence 386 of the ZJU-MoCap dataset~\cite{Peng_2021_CVPR}.}
    \label{tab:nvs_seg_res}
    \begin{tabular}{lc|lc}
    \hline
    \textbf{Split.} & \textbf{mIoU} & \textbf{Split.} & \textbf{mIoU}  \\
    \hline
    \textbf{377} & 87.48 & \textbf{386} & 89.34 \\
    \textbf{387} & 88.65 &  \textbf{392} & 85.52 \\
    \textbf{393} & 83.75 & \textbf{394} & 86.65 \\
    \hline
    \end{tabular}
    
\end{table}

\begin{figure*}[h!]
    \centering
    \begin{overpic}[width=1.\linewidth]{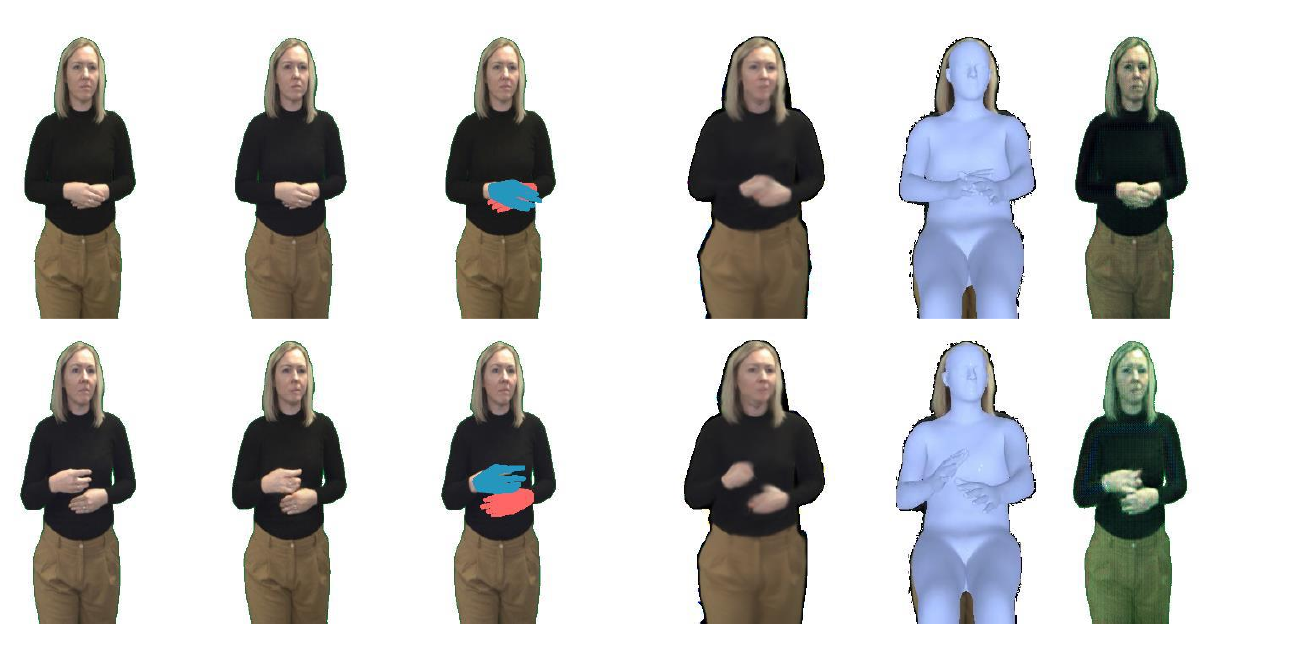}
    \put(1,0){\footnotesize Ground-truth}
    \put(21,0){\footnotesize ours}
    \put(33,0){\footnotesize NV seg. (Sec.~\textcolor{red}{3.4})}
    \put(54,0){\footnotesize GauHuman~\cite{hu2023gauhuman}}
    \put(70,0){\footnotesize EasyMocap~\cite{peng2021neural}}
    \put(84,0){\footnotesize AnonySign~\cite{Saunders_2021_FG}}
    
  \end{overpic}
    \caption{\textbf{Sign Language Production.} Qualitative results comparing our method with two state-of-the-art solid approaches. We overlay the rendering of the hand-fitted from the triangulated skeleton in our Novel-view segmentation module (Sec.~\textcolor{red}{3.4}) and the fitted SMPL parameters with the EasyMocap toolbox~\cite{peng2021neural}.}
    \label{fig:slp_soa_comp}
\end{figure*}

\subsection{Application: Sign Language Production} We apply our GaussianGAN to SLP and demonstrate the effectiveness of our approach beyond animatable avatars. First, we record a sequence of signs using a capture system composed of six synchronised cameras. 
We then triangulate the estimated skeleton, to capture the manual (hands) and non-manual (mouth) features of signs~\cite{Saunders_2022_CVPR}. The 3D facial landmarks are projected directly onto the image plane of the target view. We also implement the novel-view segmentation module using the method described in~\cite{IvashechkinMaksymTHAB}, since the segmentation of the hands provides enough information to capture the manuals.

A motivation for using the SMPL alternative approach is the lack of fine-grained detail in high-motion scenarios. Sign language is one such application where an incorrect prediction in the SMPL fitting results to a completely incorrect sign. This can be problematic as the entire meaning of the conversation changes. Therefore, we believe that a 3D human pose could be a better alternative in such cases.

We captured a sign language dataset with challenging hand interlacement to test our observation. We compare our GaussianGAN with a state-of-the-art SLP approach~\cite{Saunders_2021_FG} and an SMPL-based Gaussian splatting approach (GauHuman~\cite{hu2023gauhuman}). The results from Tab.~\ref{tab:ablation_SLP} show that our method beats the state-of-the-art approaches in PSNR and LPIPS by a large margin. This confirms that our approach can be generalised to problems other than animatable avatars.

We train GaussianGAN on the captured data and test its performance in synthesising novel poses. Fig.~\ref{fig:slp_soa_comp} shows a pose sequence of the synthesis images. Our GaussianGAN can capture photorealistic signers with manuals and non-manuals.

When using AnonySign~\cite{Saunders_2021_FG}, the model struggles to maintain a consistent style of images and noticeable artefacts. We can maintain a consistent style while generating crisp images thanks to our rendered novel view feature $\mathbf{F}$ and our segmentation module $\mathbf{S}$.

\begin{table}[t!]
    \footnotesize
    \centering
    \caption{\textbf{SLP.} Quantitative results comparing our method with two state-of-the-art approaches in sign language production.}
    \label{tab:ablation_SLP}
    \begin{tabular}{lccc}
    \hline
    \textbf{Method} & \textbf{PSNR} & \textbf{SSIM} & \textbf{LPIPS}  \\
    \hline
    AnonySign~\cite{Saunders_2021_FG} & 30.26 &  0.93 & 0.11 \\ 
    GauHuman~\cite{hu2023gauhuman} & 30.09 &  0.97 & 0.03 \\ \hdashline
    ours & 36.00 & 0.98 & 0.01 \\
    \hline
    \end{tabular}
    
\end{table}

\section{Conclusion} \label{sec:conc}
This paper presented GaussianGAN, a photorealistic, controllable human avatar in real-time. GaussianGAN accepts novel-view modalities and synthesises a photorealistic image of a human in a specific pose. First, we introduced a novel cloud initialisation strategy for Gaussian splatting, which uses the human skeleton prior to build a dense point cloud around the body's limbs. We carefully design our UNet generator model to output sharper images with high contrast for better visual quality through multi-scale discriminator loss. Our method finds its application in sign language production. Our GaussianGAN is the first SLP system to use neural rendering, which opens up the possibility of a more immersive experience for the deaf community.

We appreciate that our approach has its pitfalls, especially the training time. Our model inherits the problem of training GAN generators in general. This is still an open question, because neither transformer-based generators~\cite{Kwonjoon_2022_ICLR} nor diffusion models~\cite{Rombach_2022_CVPR} are fast to train. 

\clearpage

\section*{Ethical Impact Statement}
This work aims to improve the creation of photorealistic and controllable human avatars that offer both high visual quality and real-time performance. However, the ability to generate highly realistic avatars in real time could be misused for fraudulent purposes, such as creating deepfakes or other forms of digital impersonation that could violate individual privacy and consent.

\section*{Acknowledgement} This work was supported by the SNSF project ‘SMILE II’ (CRSII5 193686), the Innosuisse IICT Flagship (PFFS-21-47), EPSRC grant APP24554 (SignGPT-EP/Z535370/1) and through funding from Google.com via the AI for Global Goals scheme. This work reflects only the author’s views and the funders are not responsible for any use that may be made of the information it contains.

{\small
\bibliographystyle{ieee}
\bibliography{refs}
}

\end{document}